\documentclass[11pt,letterpaper]{article}
\usepackage{emnlp2017}
\usepackage{times}
\usepackage{latexsym}
\usepackage{xspace}
\usepackage{multicol}
\usepackage{graphicx}
\usepackage{amsmath}

\usepackage{amsmath}
\usepackage[colorinlistoftodos]{todonotes}
\usepackage{amsfonts,amsmath,amssymb,amsthm}
\usepackage{verbatim,float,url,enumerate}
\usepackage{natbib}
\usepackage{listings}
\usepackage{mathrsfs}
\usepackage{subcaption}
\usepackage{framed}

\theoremstyle{remark}

\theoremstyle{definition}

\emnlpfinalcopy

\def\emnlppaperid{37}

\def\CNN{{ CNN/Daily Mail}\xspace}
\def\cbt{{ Children’s Book Test}\xspace}
\def\bt{{ Book Test}\xspace}

\def\wdw{{Who-Did-What} \xspace}

\def\name{RACE\xspace}
\def\namem{RACE-M\xspace}
\def\nameh{RACE-H\xspace}
\def\dag{$\dagger$\xspace}

\title{RACE: Large-scale ReAding Comprehension Dataset From Examinations}

\author{Guokun Lai\thanks{* indicates equal contribution} \and Qizhe Xie\footnotemark[1] \and Hanxiao Liu \and Yiming Yang \and Eduard Hovy \\
\{guokun, qzxie, hanxiaol, yiming, hovy\}@cs.cmu.edu \\
Language Technologies Institute \\ Carnegie Mellon University\\ Pittsburgh, PA 15213}

\date{}

\begin{document}

\maketitle

\begin{abstract}
We present \name,
a new dataset for benchmark evaluation of methods in the reading comprehension task. Collected from the English exams for middle and high school Chinese students in the age range between 12 to 18, \name consists of near 28,000 passages and near 100,000 questions generated by human experts (English instructors), and covers a variety of topics which are carefully designed for evaluating the students' ability in  understanding and reasoning. In particular, the proportion of questions that requires reasoning is much larger in RACE than that in other benchmark datasets for reading comprehension,
and there is a significant gap between the performance of the state-of-the-art models (43\%) and the ceiling human performance (95\%). We hope this new dataset can serve as a valuable resource for research and evaluation in machine comprehension. The dataset is freely available at \url{http://www.cs.cmu.edu/~glai1/data/race/} and the code is available at \url{https://github.com/qizhex/RACE_AR_baselines} 
\end{abstract}

\section{Introduction}
\label{sec:intro}

Constructing an intelligence agent capable of understanding text as people is the major challenge of NLP research.
With recent advances in deep learning techniques,
it seems possible to achieve human-level performance in certain language understanding tasks,
and a surge of effort has been devoted to the machine comprehension task
where people aim to construct a system with the ability to answer questions related to a document that it has to comprehend \cite{chen2016thorough,kadlec2016text,dhingra2016gated, yang2017semi}. 

Towards this goal, several large-scale datasets \cite{rajpurkar2016squad, onishi2016did, hill2015goldilocks, trischler2016newsqa,hermann2015teaching} have been proposed,
which allow researchers to train deep learning systems and obtain results comparable to the human performance.
While having a suitable dataset is crucial for evaluating the system's true ability in reading comprehension,
the existing datasets suffer several critical limitations. Firstly, in all datasets, the candidate options are directly extracted from the context (as a single entity or a text span),
which leads to the fact that lots of questions can be solved trivially via word-based search and context-matching without deeper reasoning; this constrains the types of questions as well. 
Secondly, answers and questions of most datasets are either crowd-sourced or automatically-generated,
bringing a significant amount of noises in the datasets and limits the ceiling performance by domain experts, such as 82\% for \cbt and 84\% for Who-did-What. 
Yet another issue in existing datasets is that the topic coverages are often biased due to the specific ways that the data were initially collected,
making it hard to evaluate the ability of systems in text comprehension over a broader range of topics.

To address the aforementioned limitations,
we constructed a new dataset by collecting a large set of questions, answers and associated passages in the English exams for middle-school and high-school Chinese students within the 12--18 age range. Those exams were designed by domain experts (instructors) for evaluating the reading comprehension ability of students, with ensured quality and broad topic coverage.
Furthermore, the answers by machines or by humans can be objectively graded for evaluation and comparison using the same evaluation metrics.
Although efforts have been made with a similar motivation, including the MCTest dataset created by \cite{richardson2013mctest} (containing 500 passages and 2000 questions) and several others \cite{penas2014overview, rodrigo2015overview, khashabi2016question, shibuki2014overview},
the usefulness of those datasets is significantly restricted due to their small sizes, especially not suitable for training powerful deep neural networks whose success relies on the availability of relatively large training sets.

Our new dataset, namely \name,
consists of 27,933 passages and 97,687 questions.
After reading each passage, each student is asked to answer several questions where each question is provided with four candidate answers -- only one of them is correct .
Unlike existing datasets, both the questions and candidate answers in RACE are not restricted to be the text spans in the original passage; instead, they can be described in any words.
A sample from our dataset is presented in Table \ref{tab:sample}.

Our latter analysis shows that correctly answering a large portion of questions in \name requires the ability of reasoning,
the most important feature as a machine comprehension dataset \cite{chen2016thorough}.
RACE also offers two important subdivisions of the reasoning types in its questions,
namely passage summarization and attitude analysis,
which have not been introduced by the any of the existing large-scale datasets to our knowledge. 

In addition,
compared to other existing datasets where passages are either domain-specific or of a single fixed style
(namely news stories for CNN/Daily Mail, NEWSQA and Who-did-What,
fiction stories for Children's Book Test and Book Test,
and Wikipedia articles for SQUAD),
passages in RACE almost cover all types of human articles, 
such as news, stories, ads, biography, philosophy, etc., in a variety of styles.
This comprehensiveness of topic/style coverage makes RACE a desirable resource
for evaluating the reading comprehension ability of machine learning systems in general. 

The advantages of our proposed dataset over existing large datasets in machine reading comprehension can be summarized as follows:

\begin{itemize}
    \item All questions and candidate options are generated by human experts,
        which are intentionally designed to test human agent's ability in reading comprehension.
        This makes RACE a relatively accurate indicator for reflecting the text comprehension ability of machine learning systems under human judge. 

    \item The questions are substantially more difficult than those in existing datasets, in terms of the large portion of questions involving reasoning.
        At the meantime, it is also sufficiently large to support the training of deep learning models.

    \item Unlike existing large-scale datasets, candidate options in \name are human generated sentences
        which may not appear in the original passage.
        This makes the task more challenging and allows
        a rich type of questions such as passage summarization and attitude analysis.

    \item Broad coverage in various domains and writing styles:
        a desirable property for evaluating generic (in contrast to domain/style-specific) comprehension ability of learning models.

\end{itemize}

\begin{table*}[th]
	\centering
	\fbox{\begin{minipage}[t]{450pt}
{\footnotesize
{\bf Passage:}

In a small village in England about 150 years ago, a mail coach was standing on the street. It didn't come to that village often. People had to pay a lot to get a letter. The person who sent the letter didn't have to pay the postage, while the receiver had to. 

``Here's a letter for Miss Alice Brown," said the mailman.

`` I'm  Alice Brown," a girl of about 18 said in a low voice.

Alice looked at the envelope for a minute, and then handed it back to the mailman.

``I'm sorry I can't take it, I don't have enough money to pay it", she said.

A gentleman standing around were very sorry for her. Then he came up and paid the postage for her.

When the gentleman gave the letter to her, she said with a smile, `` Thank you very much, This letter is from Tom. I'm going to marry him. He went to London to look for work. I've waited a long time for this letter, but now I don't need it, there is nothing in it."

``Really? How do you know that?" the gentleman said in surprise.

``He told me that he would put some signs on the envelope. Look, sir, this cross in the corner means that he is well and this circle means he has found work. That's good news."

The gentleman was Sir Rowland Hill. He didn't forgot Alice and her letter.

``The postage to be paid by the receiver has to be changed," he said to himself and had a good plan.

``The postage has to be much lower, what about a penny? And the person who sends the letter pays the postage. He has to buy a stamp and put it on the envelope." he said .
The government accepted his plan. Then the first stamp was put out in 1840. It was called the ``Penny Black". It had a picture of the Queen on it.

\vspace{1ex}

{\bf Questions:}
\begin{multicols}{2}
1): The first postage stamp was made \_.

A. in England
B. in America
C. by Alice
D. in 1910

\vspace{1ex}
2): The girl handed the letter back to the mailman because \_ .

A. she didn't know whose letter it was

B. she had no money to pay the postage

C. she received the letter but she didn't want to open it

D. she had already known what was written in the letter

\vspace{1ex}
3): We can know from Alice's words that \_ .

A. Tom had told her what the signs meant before leaving

B. Alice was clever and could guess the meaning of the signs

C. Alice had put the signs on the envelope herself

D. Tom had put the signs as Alice had told him to

\columnbreak

4): The idea of using stamps was thought of by \_ .

A. the government

B. Sir Rowland Hill

C. Alice Brown

D. Tom

\vspace{1ex}

5): From the passage we know the high postage made \_ .

A. people never send each other letters

B. lovers almost lose every touch with each other

C. people try their best to avoid paying it

D. receivers refuse to pay the coming letters

\vspace{1ex}

{\bf Answer:} ADABC
\end{multicols}
}
\end{minipage}}
\caption{Sample reading comprehension problems from our dataset.}
\label{tab:sample}
\end{table*}

\section{Related Work}
\label{sec:related}
In this section, we briefly outline existing datasets for the machine reading comprehension task,
including their strengths and weaknesses.

\subsection{MCTest}

MCTest \cite{richardson2013mctest} is a popular dataset for question answering in the same format as \name,
where each question is associated with four candidate answers with a single correct answer. 
Although questions in MCTest are of high-quality
ensured by careful examinations through crowdsourcing,
it contains only 500 stores and 2000 questions,
which substantially restricts its usage in training advanced machine comprehension models.
Moreover, while MCTest is designed for 7 years old children,
\name is constructed for middle and high school students at 12--18 years old
hence is more complicated and requires stronger reasoning skills. 
In other words, \name can be viewed as a larger and more difficult version of the MCTest dataset.

\subsection{Cloze-style datasets}
The past few years have witnessed several large-scale cloze-style datasets~\citep{hermann2015teaching, hill2015goldilocks, bajgar2016embracing, onishi2016did}, whose questions are formulated by obliterating a word or an entity in a sentence. 

\CNN \cite{hermann2015teaching} are the largest machine comprehension datasets with 1.4M questions.
However, both require limited reasoning ability \cite{chen2016thorough}.
In fact, the best machine performance obtained by researchers \cite{chen2016thorough,dhingra2016gated} is close to human's performance on \CNN. 

\cbt(CBT) \cite{hill2015goldilocks} and \bt (BT) \cite{bajgar2016embracing} are constructed in a similar manner.
Each passage in CBT consist of $20$ contiguous sentences extracted from children's books and the next (21st) sentence is used to make the question.
The main difference between the two datasets is the size of BT being $60$ times larger.
Machine comprehension models have also matched human performance on CBT \cite{bajgar2016embracing}.

Who Did What (WDW) \cite{onishi2016did} is yet another cloze-style dataset constructed from the LDC English Gigaword newswire corpus. The authors generate passages and questions by picking two news articles describing the same event, using one as the passage and the other as the question. 

High noise is inevitable in cloze-style datasets due to their automatic generation process, which is reflected in the human performance on these datasets: $82\%$ for CBT and $84\%$ for WDW. 

\subsection{Datasets with Span-based Answers}
In datasets such as SQUAD~\citep{rajpurkar2016squad}, NEWSQA~\citep{trischler2016newsqa} MS MARCO~\citep{nguyen2016ms} and recently proposed TriviaQA~\citep{joshi2017triviaqa}.
the answer to each question is in the form of a text span in the article. Articles of SQUAD, NEWSQA and MS MARCO come from Wikipedia, CNN news and the Bing search engine respectively. The answer to a certain question may not be unique and could be multiple spans. Instead of evaluating the accuracy, researchers need to use F1 score, BLEU~\citep{papineni2002bleu} or ROUGE~\citep{lin2003automatic} as metrics, which measure the overlap between the prediction and ground truth answers since the questions come without candidate spans.

Datasets with span-based answers are challenging as the space of possible spans is usually large. 
However,
restricting answers to be text spans in the context passage may be unrealistic and more importantly, may not be intuitive even for humans,
indicated by the suffered human performance of 80.3\% on SQUAD (or 65\% claimed by \citet{trischler2016newsqa}) and 46.5\% on NEWSQA.
In other words, the format of span-based answers may not necessarily be a good examination of reading comprehension of machines whose aim is to approach the comprehension ability of \emph{humans}. 

\subsection{Datasets from Examinations}
There have been several datasets extracted from examinations,
aiming at evaluating systems under the same conditions as how humans are evaluated in schools.
E.g.,
the AI2 Elementary School Science Questions dataset~\citep{khashabi2016question} contains 1080 questions for students in elementary schools; 
NTCIR QA Lab~\citep{shibuki2014overview} evaluates systems by the task of solving real-world university entrance exam questions;
The Entrance Exams task at CLEF QA Track~\citep{penas2014overview, rodrigo2015overview} evaluates the system's reading comprehension ability. 
However, data provided in these existing tasks are far from sufficient for the training of advanced data-driven machine reading models,
partially due to the expensive data generation process by human experts.

To the best of our knowledge, \name is the first \emph{large-scale} dataset of this type,
where questions are created based on exams designed to evaluate human performance in reading comprehension.

\section{Data Analysis}
\label{sec:analysis}

\begin{table*}[ht]
\centering
\begin{tabular}{l|ccc|ccc|cccc}
\hline
Dataset & \multicolumn{3}{c|}{\namem} & \multicolumn{3}{c|}{\nameh} & \multicolumn{4}{c}{\name} \\ \hline
Subset                 & Train & Dev & Test & Train & Dev & Test & Train & Dev & Test & All    \\ \hline
\# passages        & 6,409    & 368   & 362    & 18,728   & 1,021       & 1,045  &25,137 &1,389& 1,407 & 27,933  \\
\# questions       & 25,421   & 1,436  & 1,436   & 62,445   & 3,451       & 3,498  &87,866 &4,887 &4,934 & 97,687  \\
\hline
\end{tabular}
\caption{The separation of the training, development and test sets of \namem,\nameh and \name}
\label{tab:subset_split}
\end{table*}

\begin{table}[ht]
\centering
\begin{tabular}{l|c|c|c} \hline
Dataset & \namem & \nameh & \name \\ \hline
Passage Len  & 231.1               & 353.1               & 321.9  \\
Question Len & 9.0                 & 10.4               & 10.0    \\
Option Len & 3.9                 & 5.8                 & 5.3    \\
Vocab size        & 32,811               & 125,120               & 136,629 \\
\hline
\end{tabular}
\caption{Statistics of \name where Len denotes length and Vocab denotes Vocabulary.}
\label{tab:sta}
\end{table}

\iffalse

\begin{table*}[ht]
\centering
\begin{tabular}{l|ccc|ccc|ccc|}
\hline
Dataset & \multicolumn{3}{c|}{\namem} & \multicolumn{3}{c|}{\nameh} & \name \\ \hline
Subset                 & Train & Dev & Test & Train & Dev & Test &    \\ \hline
\# passages        & 6,456    & 373   & 362    & 18,859   & 1,028       & 1,052  & 28,130  \\
\# questions       & 25,634   & 1,460  & 1,436   & 62,900   & 3,474       & 3,528  & 98,432  \\
Avg \# words per passage  & \multicolumn{3}{c|}{202.3}               & \multicolumn{3}{c|}{317.4}               & 285.2  \\
Avg \# words per question & \multicolumn{3}{c|}{8.8}                 & \multicolumn{3}{c|}{10.1}                & 9.7    \\
Avg \# words per option   & \multicolumn{3}{c|}{4.6}                 & \multicolumn{3}{c|}{6.4}                 & 5.7    \\
Vocabulary Size        & \multicolumn{3}{c|}{28,599}               & \multicolumn{3}{c|}{93,332}               & 99,323 \\
\hline
\end{tabular}
\caption{Statistics of \name}
\label{tab:sta}
\end{table*}

\fi

\begin{table*}[ht]
\centering
\begin{tabular}{l|cccccc}
\hline
Dataset       & \namem  & \nameh  & \name   & CNN & SQUAD   & NEWSQA  \\
\hline
Word Matching & 29.4\% & 11.3\% & 15.8\% & 13.0\%$^\dagger$ & 39.8\%* & 32.7\%* \\
Paraphrasing   & 14.8\% & 20.6\% & 19.2\% & 41.0\%$^\dagger$ & 34.3\%* & 27.0\%* \\
Single-Sentence Reasoning & 31.3\% & 34.1\% &33.4\% & 19.0\%$^\dagger$& 8.6\%* & 13.2\%*\\
Multi-Sentence Reasoning  & 22.6\% & 26.9\% & 25.8\% & 2.0\%$^\dagger$& 11.9\%* & 20.7\%* \\
Ambiguous/Insufficient     & 1.8\%  & 7.1\%  & 5.8\%  & 25.0\%$^\dagger$& 5.4\%*  & 6.4\%*  \\
\hline
\end{tabular}
\caption{Statistic information about Reasoning type in different datasets. * denotes the numbers coming from \cite{trischler2016newsqa} based on 1000 samples per dataset, and numbers with $\dagger$ come from \cite{chen2016thorough}.}
\label{tab:reason}
\end{table*}
In this section, we study the nature of questions covered in RACE at a detailed level. Specifically, we present the dataset statistics in Section \ref{sec:sta},
and then analyze different reasoning/question types in RACE in the remaining subsections.

\subsection{Dataset Statistics}
\label{sec:sta}
As mentioned in section \ref{sec:intro}, \name is collected from English examinations designed for 12--15 year-old middle school students, and 15--18 year-old high school students in China. To distinguish the two subgroups with drastic difficulty gap, \namem denotes the middle school examinations and \nameh denotes high school examinations. We split 5\% data as the development set and 5\% as the test set for \namem and \nameh respectively. The number of samples in each set is shown in Table \ref{tab:subset_split}. The statistics for \namem and \nameh is summarized in Table \ref{tab:sta}.
We can find that the length of the passages and the vocabulary size in the \nameh are much larger than that of the \namem, an evidence of the higher difficulty of high school examinations. 

However, notice that since the articles and questions are selected and designed to test Chinese students learning English as a foreign language, the vocabulary size and the complexity of the language constructs are simpler than news articles and Wikipedia articles in other QA datasets. 

\subsection{Reasoning Types of the Questions}
\label{sec:reason}

To get a comprehensive picture about the reasoning difficulty requirement of \name, we conduct human annotations of questions types.
Following \citet{chen2016thorough,trischler2016newsqa}, we stratify the questions into five classes as follows with ascending order of difficulty:

\begin{itemize}
\item Word matching: The question exactly matches a span in the article. The answer is self-evident. 

\item Paraphrasing: The question is entailed or paraphrased by exactly one sentence in the passage. The answer can be extracted within the sentence. 

\item Single-sentence reasoning: The answer could be inferred from a single sentence of the article by recognizing incomplete information or conceptual overlap. 

\item Multi-sentence reasoning: The answer must be inferred from synthesizing information distributed across multiple sentences. 

\item Insufficient/Ambiguous: The question has no answer or the answer is not unique based on the given passage.  
\end{itemize}

We refer readers to \cite{chen2016thorough,trischler2016newsqa} for examples of each category. 

To obtain the proportion of different question types, we sample $100$ passages from RACE ($50$ from \namem and $50$ from \nameh), all of which have $5$ questions hence there are $500$ questions in total. We put the passages on Amazon Mechanical Turk\footnote{https://www.mturk.com/mturk/welcome}, and a Hit is generated by a passage with 5 questions. Each question is labeled by two crowdworkers. We require the turkers to both answer the questions and label the reasoning type. We pay \$0.70 and \$1.00 per passage in \namem and \nameh respectively, and restrict the access to master turkers only. Finally, we get 1000 labels for the 500 questions. 

The statistics about the reasoning type is summarized in Table \ref{tab:reason}. The higher difficulty level of \name is justified by its higher ratio of reasoning questions in comparison to CNN, SQUAD and NEWSQA. Specifically,
$59.2\%$ questions of \name are either in the category of single-sentence reasoning or in the category of multi-sentence reasoning, while the ratio is $21\%$, $20.5\%$ and $33.9\%$ for CNN, SQUAD and NEWSQA respectively.
Also notice that the ratio of word matching questions on \name is only $15.8\%$, the lowest among several categories.
In addition, questions in \nameh are more complex than questions in \namem since \namem has more word matching questions and fewer reasoning questions. 

\subsection{Subdividing Reasoning Types}
\label{sec:sub_reason}

To better understand our dataset and facilitate future research, we list the subdivisions of questions under the reasoning category. 
We find the most frequent reasoning subdivisions include: detail reasoning, whole-picture understanding, passage summarization, attitude analysis and world knowledge.  One question may fall into multiple divisions. Definition of these subdivisions and their associated examples are as follows:

1. Detail reasoning: to answer the question, the agent should be clear about the details of the passage. The answer appears in the passage but it cannot be found by simply matching the question with the passage.
For example, Question $1$ in the sample passage falls into this category.

2. Whole-picture reasoning: the agent needs to understand the whole picture of the story to obtain the correct answer. For example, to answer the Question $2$ in the sample passage, the agent is required to comprehend the entire story.  

3. Passage summarization: The question requires the agent to select the best summarization of the passage among four candidate summarizations. A typical question of this type is ``The main idea of this passage is \underline{\hspace{1em}}.''.
An example question can be found in Appendix \ref{sec:ex_q_summerization}.

4. Attitude analysis: The question asks about the opinions/attitudes of the author or a character in the story towards somebody or something, e.g., 

\begin{framed}

{\small
\begin{itemize}

\item \emph{Evidence}: ``\ldots Many people optimistically thought industry awards for better equipment would stimulate the production of quieter appliances. It was even suggested that noise from building sites could be alleviated \ldots''

\item \emph{Question}: What was the author's attitude towards the industry awards for quieter?
 
\item \emph{Options}: A.suspicious \enskip B.positive \enskip
C.enthusiastic \enskip D.indifferent

\end{itemize}
}
 
\end{framed}

5. World knowledge: Certain external knowledge is needed. Most frequent questions under this category involve simple arithmetic.
{\small
\begin{framed}

\begin{itemize}
\item \emph{Evidence}: ``The park is open from 8 am to 5 pm.''
\item \emph{Question}: The park is open for \underline{\hspace{1em}}  hours a day.
\item \emph{Options}: A.eight \hspace{0.2cm}B.nine \hspace{0.2cm}C.ten \hspace{0.2cm}D.eleven
\end{itemize}

\end{framed}
 }

To the best of our knowledge, questions like passage summarization and attitude analysis have not been introduced by any of the existing large-scale machine comprehension datasets. Both are crucial components in evaluating humans' reading comprehension abilities.

\section{Collection Methodology}
\label{sec:collection}

We collected the raw data from three large  free public websites in China\footnote{We checked that our dataset does not include example questions of exams with copyright, such as SSAT, SAT, TOEFL and GRE.}, where the reading comprehension problems are extracted from English examinations designed by teachers in China. The data before cleaning contains 137,918 passages and 519,878 questions in total, where there are 38,159 passages with 156,782 questions in the middle school group, and 99,759 passages with 363,096 questions in the high school group. 

The following filtering steps are conducted to clean the raw data. Firstly, we remove all problems and questions that do not have the same format as our problem setting, e.g., a question would be removed if the number of its options is not four. 
Secondly, we filter all articles and questions that are not self-contained based on the text information, i.e. we remove the articles and questions containing images or tables. We also remove all questions containing keywords ``underlined" or ``paragraph", since it is difficult to reproduce the effect of underlines and the paragraph segment information.
Thirdly, we remove all duplicated articles. 

On one of the websites (xkw.com), the answers are stored as images. 
We used two standard OCR programs tesseract \footnote{https://github.com/tesseract-ocr} and ABBYY FineReader \footnote{https://www.abbyy.com/FineReader} to process the images. 
We remove all the answers that two software disagree.
The OCR task is easy since we only need to recognize printed alphabet A, B, C, D with a standard font. Finally, we get the cleaned dataset \name, with 27,933 passages and 97,687 questions. 

\section{Experiments}
\label{sec:experiment}

\begin{table*}[ht]
\centering
\resizebox{\textwidth}{!}{
\begin{tabular}{l|@{\hskip3pt}c@{\hskip3pt}c@{\hskip3pt}c@{\hskip3pt}c@{\hskip3pt}c@{\hskip3pt}c@{\hskip3pt}c@{\hskip3pt}c@{\hskip3pt}c}
\hline
               & RACE-M & RACE-H & RACE   & MCTest & CNN   & DM & CBT-N & CBT-C & WDW   \\
\hline
Random         & 24.6      & 25.0      & 24.9      & 24.8      &0.06     & 0.06            & 10.6      & 10.2   & 32.0$^\dagger$ \\
Sliding Window & 37.3     & 30.4     & 32.2     & 51.5$^\dagger$ & 24.8     & 30.8            & 16.8$^\dagger$  & 19.6$^\dagger$  & 48.0$^\dagger$    \\
Stanford AR    & 44.2     & 43.0     & 43.3     & --       &73.6$^\dagger$&    76.6$^\dagger$      &  --     & --      & 64.0$^\dagger$ \\
GA      & 43.7& 44.2 &44.1 & -- & 77.9$^\dagger$ & 80.9$^\dagger$ & 70.1$^\dagger$ & 67.3$^\dagger$ & 71.2$^\dagger$ \\
\hline
Turkers         & 85.1 & 69.4 & 73.3 & -- & -- & -- & -- & -- & --  \\
Ceiling Performance          & 95.4 & 94.2 & 94.5 & -- & -- & -- & 81.6$^\dagger$ & 81.6$^\dagger$ & 84$^\dagger$ \\
\hline
\end{tabular}
}
\caption{Accuracy of models and human on the each dataset, where \dag denotes the results coming from previous publications. DM denotes Daily Mail and WDW denotes \wdw. }
\label{tab:result}
\end{table*}

\begin{figure*}[!ht]
\centering
\begin{subfigure}{.45\textwidth}
  \includegraphics[width=\linewidth]{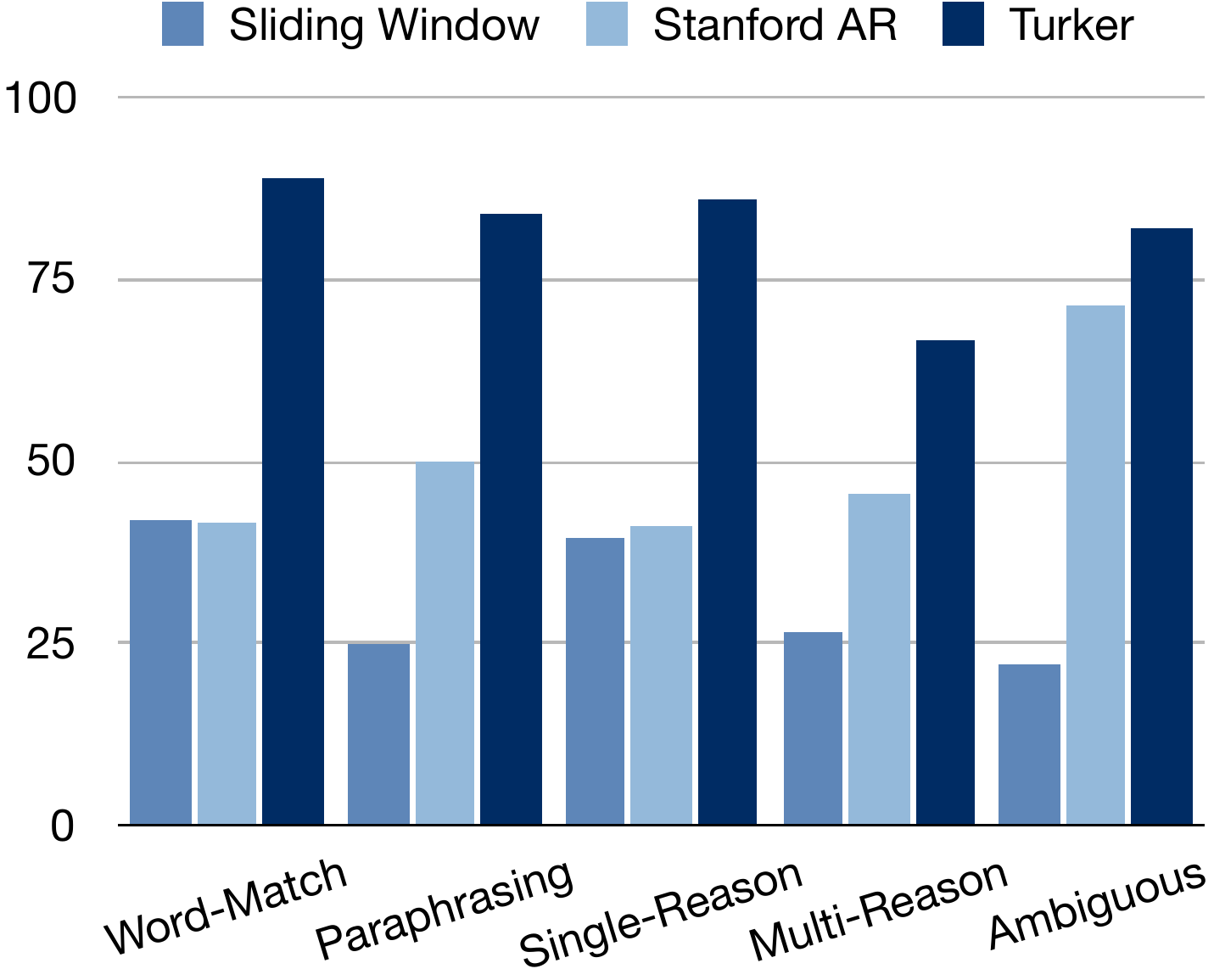}
  \caption{\namem}
\end{subfigure}
\begin{subfigure}{.45\textwidth}
  \includegraphics[width=\linewidth]{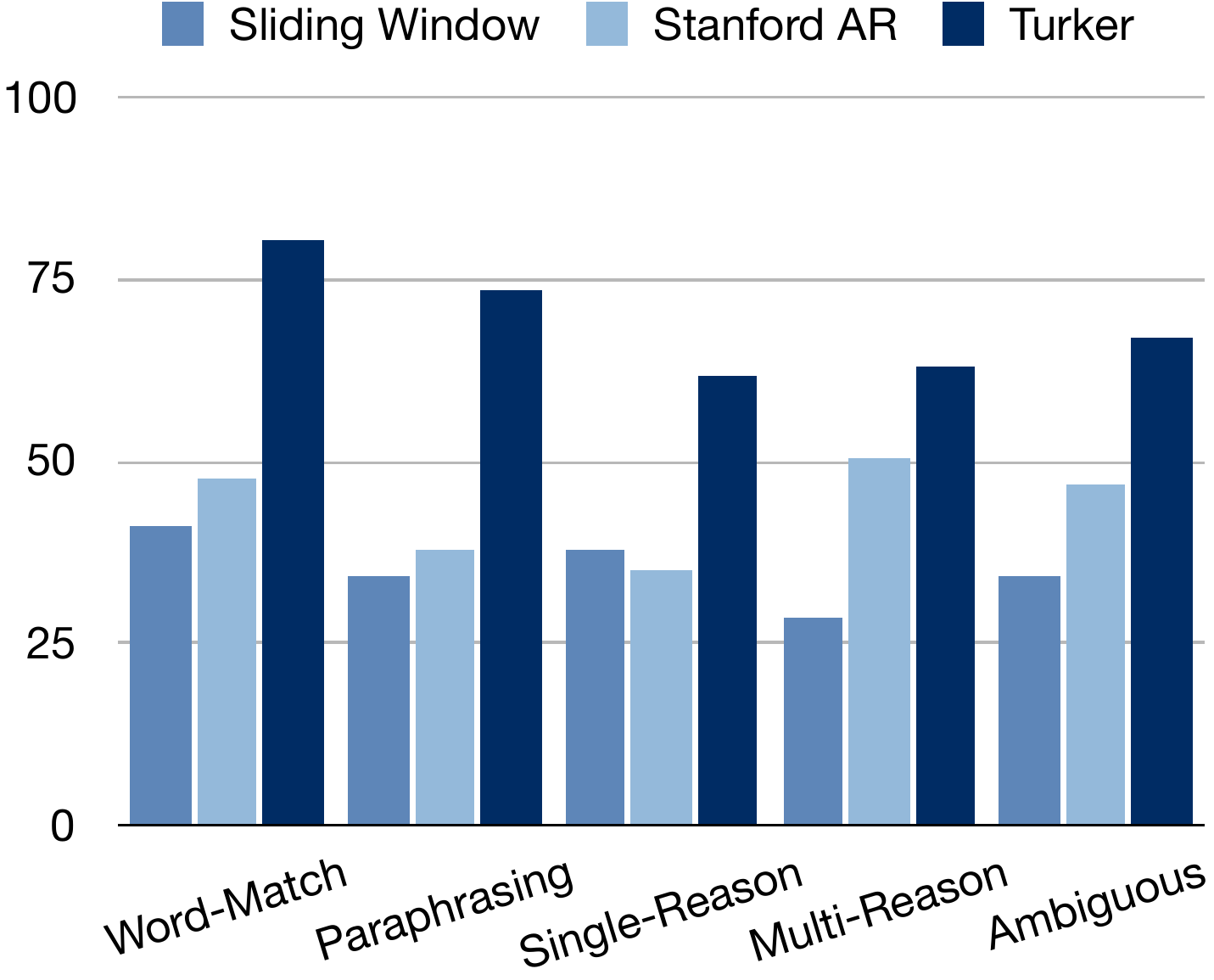}
  \caption{\nameh}
\end{subfigure}
\caption{Test accuracy of different baselines on each question type category introduced in Section \ref{sec:reason}, where Word-Match, Single-Reason, Multi-Reason and Ambiguous are the abbreviations for Word matching, Single-sentence Reasoning, Multi-sentence Reasoning and Insufficient/Ambiguous respectively. } 

\label{fig:resultgroup}
\end{figure*}

In this section, we compare the performance of several state-of-the-art reading comprehension models with human performance. We use accuracy as the metric to evaluate different models.

\subsection{Methods for Comparison}
\paragraph{Sliding Window Algorithm}
Firstly, we build the rule-based baseline introduced by \citet{richardson2013mctest}. It chooses the answer having the highest matching score. Specifically, it first concatenates the question and the answer and then calculates the TF-IDF style matching score between the concatenated sentence with every window (a span of text) of the article. The window size is decided by the model performance in the training and dev sets. 

\paragraph{Stanford Attentive Reader}
Stanford Attentive Reader (Stanford AR)~\citep{chen2016thorough} is a strong model that achieves state-of-the-art results on \CNN. Moreover, the authors claim that their model has nearly reached the ceiling performance on these two datasets. 

Suppose that the triple of passage, question and options is denoted by $(p, q, o_{1, \cdots, 4})$. We first employ bidirectional GRUs to encode $p$ and $q$ respectively into $h^p_{1}, h^p_{2}, \dots, h^p_{n}$ and $h^q$. Then we summarize the most relevant part of the passage into $s^{p}$ with an attention model. Following~\citet{chen2016thorough}, we adopt a bilinear attention form. Specifically, 

\begin{equation}
\begin{aligned}
\alpha_i &= \mathrm{Softmax}_i((h^p_{i})^T W_1 h^q)\\
s^p &= \sum_i \alpha_i h^p_{i}
\end{aligned}
\end{equation}

Similarly, we use bidirectional GRUs to encode option $o_i$ into a vector $h^{o_i}$. Finally, we compute the matching score between the $i$-th option $(i=1, \cdots, 4)$ and the summarized passage using a bilinear attention. We pass the scores through softmax to get a probability distribution. Specifically, the probability of option $i$ being the right answer is calculated as
\begin{equation}
p_i = \mathrm{Softmax}_i(h^{o_i} W_2 s^d)
\label{eq:output_prob}
\end{equation}

\paragraph{Gated-Attention Reader} Gated AR~\citep{dhingra2016gated} is the state-of-the-art model on multiple datasets. To build query-specific representations of tokens in the document, it employs an attention mechanism to model multiplicative interactions between the query embedding and the document representation. With a multi-hop architecture, GA also enables a model to scan the document and the question iteratively for multiple passes. In other words, the multi-hop structure makes it possible for the reader to refine token representations iteratively and the attention mechanism find the most relevant part of the document. We refer readers to ~\citep{dhingra2016gated} for more details. 

After obtaining a query specific document representation $s^d$, we use the same method as bilinear operation listed in Equation \ref{eq:output_prob} to get the output.

Note that our implementation slightly differs from the original GA reader. Specifically, the Attention Sum layer is not applied at the final layer and no character-level embeddings are used. 

\paragraph{Implementation Details} We follow ~\citet{chen2016thorough} in our experiment settings. The vocabulary size is set to $50k$. We choose word embedding size $d=100$ and use the $100$-dimensional Glove word embedding~\citep{pennington2014glove} as embedding initialization. GRU weights are initialized from Gaussian distribution $\mathcal{N}(0, 0.1)$. Other parameters are initialized from a uniform distribution on $(-0.01, 0.01)$. The hidden dimensionality is set to $128$ and the number of layers is set to one for both Stanford AR and GA. 
We use vanilla stochastic gradient descent (SGD) to train our models. We apply dropout on word embeddings and the gradient is clipped when the norm of the gradient is larger than $10$.
We use a grid search on validation set to choose the learning rate within $\{0.05, 0.1, 0.3, 0.5\}$ and dropout rate within $\{0.2, 0.5, 0.7\}$. The highest accuracy on validation set is obtained by setting learning rate to $0.1$ for Stanford AR and $0.3$ for GA and dropout rate to $0.5$. 
 The data of \namem and \nameh is used together to train our model and testing is performed separately. 

\subsection{Human Evaluation}
\label{sec:human_perform}

As described in section \ref{sec:reason}, a randomly sampled subset of test set has been labeled by Amazon Turkers, which contains 500 questions with half from \nameh and with the other half from \namem. The turkers' performance is 85\% for \namem and 70\% for \nameh. However, it is hard to guarantee that every turker performs the survey carefully, given the difficult and long passages of high school problems. Therefore, to obtain the ceiling human performance on \name, 
we manually labeled the proportion of valid questions. A question is valid if it is unambiguous and has a correct answer. We found that 94.5\% of the data is valid, which sets the ceiling human performance. Similarly, the ceiling performance on \namem and \nameh is 95.4\% and 94.2\% respectively.

\subsection{Main Results}
We compare models' and human ceiling performance on datasets which have the same evaluation metric with \name. The compared datasets include \name, MCTest, \CNN (CNN and DM), CBT and  WDW. 
On CBT, we report performance on two subsets where the missing token is either a common noun (CBT-C) or name entity (CBT-N) since the language models have already reached human-level performance on other types \cite{hill2015goldilocks}.  The comparison is shown in Table \ref{tab:result}.

\paragraph{Performance of Sliding Window} 
We first compare MCTest with \name using Sliding Window, where it is unable to train Stanford AR and Gated AR on MCTest's limited training data. Sliding Window achieves an accuracy of $51.5\%$ on MCTest while only $37.3\%$ on \name, meaning that to answer the questions of \name requires more reasoning than MCTest. 

The performance of sliding window on \name is not directly comparable with CBT and WDW since CBT has ten candidate answers for each question and WDW has an average of three.  Instead, we evaluate the performance improvement of sliding window on the random baseline. Larger improvement indicates more questions solvable by simple matching. On \name, Sliding Window is $28.6\%$ better than the random baseline, while the improvement is $58.5\%$, $92.2\%$ and $50\%$ for CBT-N, CBT-C and WDW.

The accuracy on \namem (37.3\%) and \nameh (30.4\%) indicates that the middle school questions are simpler based on the matching algorithm. 

\paragraph{Performance of Neural Models} We further compare the difficulty of different datasets by state-of-the-art neural models' performance. A lower performance means that more problems are unsolvable by machines. The Stanford AR and Gated AR achieve an accuracy of only $43.3\%$ and $44.1\%$ on \name while their accuracy is much higher on \CNN, \cbt and Who-Did-What. It justifies the fact that, among current large-scale machine comprehension datasets, \name is the most challenging one. 

\paragraph{Human Ceiling Performance}
The human performance is $94.5\%$ which shows our data is quite clean compared to other large-scale machine comprehension datasets.  Since we cannot enforce every turker do the test cautiously, the result shows a gap between turkers' performance and human performance. Reasonably, problems in the high school group with longer passages and more complex questions lead to more significant divergence. 
Nevertheless, the start-of-the-art models still have a large room to be improved to reach turkers' performance. The performance gap is 41\% for the middle school problems and 25\% for the high school problems. What's more, The performance of Stanford AR and GA is only less than a half of the ceiling human performance, which indicates that to match the humans' reading comprehension ability, we still have a long way to go. 

\subsection{Reason Types Analysis}
We evaluate human and models on different types of questions, shown in Figure \ref{fig:resultgroup}. Turkers do the best on word matching problems while doing the worst on reasoning problems. Sliding window performs better on word matching than problems needing reasoning or paraphrasing. Surprisingly, Stanford AR does not have a stronger performance on the word matching category than reasoning categories. A possible reason is that the proportion of data in reasoning categories is larger than that of data. Also, the candidate answers of simple matching questions may share similar word embeddings. For example, if the question is about color, it is difficult to distinguish candidate answers, ``green", ``red", ``blue" and ``yellow", in the embedding vector space. 
The similar performance on different categories also explains the reason that the performance of the neural models is close in the middle and high school groups in Table \ref{tab:result}.

\section{Conclusion}
We introduce a large, high-quality dataset for reading comprehension that is carefully designed to examine human ability on this task. 
Some desirable properties of \name include the broad coverage of domains/styles and the richness in the question format.
Most importantly,
it requires substantially more reasoning to do well on \name than on other datasets, as there is a significant gap between the performance of state-of-the-art machine comprehension models and that of the human. We hope this dataset will stimulate the development of more advanced machine comprehension models.  

\label{sec:conclusion}

\section*{Acknowledgement}
We would like to thank Graham Neubig for suggestions on the draft and Diyi Yang's help on obtaining the crowdsourced labels.

This research was supported in part by DARPA grant FA8750-12-2-0342 funded under the DEFT program.

\bibliography{emnlp2017}
\bibliographystyle{emnlp_natbib}

\clearpage

\begin{appendix}
\section{Appendix}
\subsection{Example Question of Passage Summarization}
\label{sec:ex_q_summerization}

Passage: Do you love holidays but hate gaining weight? You are not alone. Holidays are times for celebrating. Many people are worried about their weight. With proper planning, though, it is possible to keep normal weight during the holidays. The idea is to enjoy the holidays but not to eat too much. You don't have to turn away from the foods that you enjoy.

Here are some tips for preventing weight gain and maintaining physical fitness:

Don't skip meals. Before you leave home, have a small, low-fat meal or snack. This may help to avoid getting too excited before delicious foods.

Control the amount of food. Use a small plate that may encourage you to "load up". You should be most comfortable eating an amount of food about the size of your fist.

Begin with soup and fruit or vegetables. Fill up beforehand on water-based soup and raw fruit or vegetables, or drink a large glass of water before you eat to help you to feel full.

Avoid high-fat foods. Dishes that look oily or creamy may have large amount of fat. Choose lean meat . Fill your plate with salad and green vegetables. Use lemon juice instead of creamy food.

Stick to physical activity. Don't let exercise take a break during the holidays. A 20-minute walk helps to burn off extra calories.

Questions:

What is the best title of the passage?

Options:

A. How to avoid holiday feasting

B. Do's and don'ts for keeping slim and fit.

C. How to avoid weight gain over holidays.

D. Wonderful holidays, boring experiences.

\end{appendix}
\end{document}